# Enhancing Financial Inclusion and Regulatory Challenges: A Critical Analysis of Digital Banks and Alternative Lenders Through Digital Platforms, Machine Learning, and Large Language Models Integration

Luke Lee
King's College London
London, United Kingdom
luke.lee@kcl.ac.uk

**ABSTRACT:**

This paper explores the dual impact of digital banks and alternative lenders on financial inclusion and the regulatory challenges posed by their business models. It discusses the integration of digital platforms, machine learning (ML), and Large Language Models (LLMs) in enhancing financial services accessibility for underserved populations. Through a detailed analysis of operational frameworks and technological infrastructures, this research identifies key mechanisms that facilitate broader financial access and mitigate traditional barriers. Additionally, the paper addresses significant regulatory concerns involving data privacy, algorithmic bias, financial stability, and consumer protection. Employing a mixed-methods approach, which combines quantitative financial data analysis with qualitative insights from industry experts, this paper elucidates the complexities of leveraging digital technology to foster financial inclusivity. The findings underscore the necessity of evolving regulatory frameworks that harmonize innovation with comprehensive risk management. This paper concludes with policy recommendations for regulators, financial institutions, and technology providers, aiming to cultivate a more inclusive and stable financial ecosystem through prudent digital technology integration.

## 1. Introduction

In recent years, rapid technological advancements have led to a dramatic increase in the uptake of digital financial services. The COVID-19 pandemic acted as a catalyst for this increase in uptake.[1] A transformative shift has occurred within the financial sector, with the adoption of new 'enabling'[2] technologies such as artificial intelligence (AI), distributed ledger technology (DLT),[3] and machine learning. Digital banks and alternative lenders have leveraged these technologies to enhance financial inclusion. However, such technologies create new risks, resulting in fresh challenges for lawmakers. Data security, a lack of transparency, and fraud are potential harms that must be adequately offset. For instance, a laxity of requirements for due diligence and Know-Your-Client (KYC) checks on alternative lending platforms in jurisdictions without a "*specific regulatory framework*"[4] can be exploited for nefarious purposes such as money laundering and the financing of terrorism.

---

[1] McKinsey & Company, 'Global Banking Practice: Building the AI Bank of the Future' (2021) 2.

[2] Johannes Ehrentraud and Denise Garcia Ocampo and Camila Quevedo Vega, *Regulating Fintech Financing: Digital Banks and Fintech Platforms* (Bank for International Settlements, Financial Stability Institute, 2020) 4.

[3] Jagadeesha R Bhat and Salman A AlQahtani and Maziar Nekovee, 'Fintech Enablers, Use Cases, and Role of Future Internet of Things' (2023) 35 JKSUCIS 1, 87.

The financial sector is now at a critical fork in the road. In a sense, regulators of the FinTech space are wrestling with problems never before encountered. They face an uphill task of examining how current rules and frameworks can be adapted.[5] In this evolving digital environment, regulators now have to conduct a delicate balancing exercise between two competing considerations: promoting financial innovation on the one hand and ensuring the integrity of the financial system on the other.[6] From a macroscopic perspective, policymakers face the unique challenge of ensuring "*that FinTech develops in a way that maximises the opportunities and minimises the risks for society*".[7] Thus, policymakers must carefully rethink established policies (and perhaps formulate new ones) to adapt sufficiently to this evolving digital ecosystem.

This paper examines the dual impact of technologies on financial inclusion and innovation risks. It focuses on the role of digital banks and alternative lenders in enhancing financial inclusion, addresses the regulatory challenges of their business models, and discusses the balance between innovation and regulation. The conclusion will offer insights into the future regulation of these financial entities.

Before delving into the substance of this paper, it is crucial to define certain key terms. What is financial inclusion? It is the ability of people and companies to utilise beneficial and reasonably priced financial services, including transaction processing, payments, savings options, loans, and insurance, in an ethical and long-lasting manner.[8] Next, what are business models? Simply put, they are ways that companies deliver value to their clients, persuade them to purchase this value, and transform those purchases into earnings.[9] I now turn to the respective definitions of digital banks and alternative lenders.

---

[4] Johannes and Denise and Camila (n 2) 20.

[5] *Ibid* 1.

[6] *Ibid* 33.

[7] Mark Carney, 'The promise of Fintech–something new under the sun' in *Speech at Deutsche Bundesbank G20 Conference, by Bank of England Governor Mark Carney* (2017).

[8] The World Bank, 'Financial Inclusion' (*The World Bank*, 29 March 2022), <https://www.worldbank.org/en/topic/financialinclusion/overview>, accessed 2 December 2023.

[9] David J Teece, 'Business Models, Business Strategy and Innovation' (2010) 43 LRP 2-3, 172 as cited in Marko Peric and Jelena Durkin and Vanja Vitezic, 'The Constructs of a Business Model Redefined: A Half-Century Journey' (2017) 7 SO 3.

Digital banks participate in deposit insurance schemes, primarily operate online, and offer banking services through electronic channels with minimal or no physical branches.[10] They manage risks like traditional banks but leverage technology-driven models for remote service delivery.[11] Digital banks can offer their customers comprehensive banking offerings and solutions.[12]

Alternative or peer-to-peer (P2P) lenders have emerged as significant FinTech disruptors. These are lenders who engage in "*lending-related FinTech platform financing activities*."[13] Alternative lenders utilise data analytics and technology-powered frameworks to extend loans to borrowers in underserved markets.[14] These borrowers would not normally qualify for loans under traditional lending establishments' requirements.[15] Alternative lenders utilise innovative credit scoring and financing models with appealing returns and shorter terms to provide lower interest rates to borrowers than traditional incumbent banks.[16] Examples of alternative lenders are the world's first P2P-lending company, Zopa (which formerly offered P2P consumer loans),[17] SoFi, LendingClub, Prosper, and OnDeck,[18] which have disrupted the conventional lending landscape by offering P2P lending and other novel credit solutions.

Digital banks and alternative lenders were formed due to a rising trend amongst banking customers for greater accessibility, transparency, and customer-centric service. It is crucial to note that "*uses of this technology in finance are in a nascent and rapidly evolving phase (...) [and] any analysis must be necessarily preliminary...*"[19] I will focus my analysis on three technologies utilised by digital banks and alternative lenders: digital platforms, machine learning, and Large Language Models.

---

[10] Johannes and Denise and Camila (n 2) 3.

[11] *Ibid*.

[12] *Ibid* 9.

[13] *Ibid* 4.

[14] Kenneth Michlitsch, 'An Introduction to Alternative Lending' (Morgan Stanley Investment Management, Investment Insight, 2021) 1-2.

[15] *Ibid* 1.

[16] *Ibid* 2-3.

[17] Rupert Jones, 'Zopa Exits Peer-to-Peer Lending to focus on Banking' (*The Guardian*, 11 December 2021) <https://www.theguardian.com/money/2021/dec/11/zopa-peer-to-peer-lending-p2p-money>, accessed on 20 December 2023.

[18] Giulio Cornelli and others, 'Fintech and Big Tech Credit: Drivers of the Growth of Digital Lending' (2023) 148 JBF art 106742, 5.

[19] Financial Stability Board, 'Artificial Intelligence and Machine Learning in Financial Services: Market Developments and Financial Stability Implications', (2017).

## 2. Enhancing Financial Inclusion through Digital Banks and Alternative Lenders

### 2.1. Digital Platforms

Digital platforms significantly advance financial inclusion, as customers accustomed to their convenience, speed, and personalised services increasingly expect similar offerings from banks. These platforms facilitate many daily financial transactions, raising customer expectations.[20] The operational cost reductions afforded by digital platforms are noteworthy due to the network connectivity and sophisticated data functionalities that digital platforms offer.[21] As digital banks are free from the financial burden of maintaining extensive physical branch networks, they can operate with much lower overheads, giving them an economic upper hand over incumbent banking models.[22] This streamlined efficiency allows digital banks to offer more competitive rates and reduced fees, in stark contrast to the cost structures of traditional banks.[23] This opens up access to financial services to customers who would traditionally not have access to them.

One notable example of the success of digital platforms is M-PESA, a prominent provider of mobile financial services in Kenya.[24] M-PESA has revolutionised mobile money banking in Africa.[25] M-PESA enables its users to perform various financial transactions through their mobile phones without requiring a standard bank account.[26] M-PESA found massive success and unprecedented mass adoption in Kenya, with subscriber growth from approximately 52,000 to 8.6 million from 2007 to 2009.[27] Likely, the rapid acceptance of M-PESA as an alternative to traditional banking services was a case of being at 'the right place at the right time'.[28] Factors such as a lack of literacy, frustrating administrative paperwork, and a high concentration of financial services only in major urban centres were significant barriers to financial inclusion for customers in rural areas of Kenya.[29] These factors necessitated a fresh approach to banking.[30] M-PESA met this need. The same is true of a 2020 case study that confirmed the beneficial effects of FinTech services on financial inclusion in Mozambique, where 77% of Mozambique's population in rural areas lacks access to financial services.[31]

---

[20] McKinsey & Company (n 1) 2.

[21] Johannes and Denise and Camila (n 2) 9.

[22] *Ibid*.

[23] *Ibid*.

[24] Benjamin Ngugi and Matthew Pelowski and Javier Gordon Ogembo, 'M-Pesa: A Case Study of the Critical Early Adopters' Role in the Rapid Adoption of Mobile Money Banking in Kenya' (2010) 43 TEJISDC 1, 1-16.

[25] *Ibid* 1-16.

[26] *Ibid* 1.

[27] *Ibid* 1-16.

[28] *Ibid*.

[29] *Ibid* 6.

[30] *Ibid* 4.

The transition from traditional to digital banking, exemplified by the case study of M-PESA, has notably expanded banking services, reaching remote and underserved areas like rural Kenya and Mozambique. This shift has been especially impactful for populations previously marginalised by conventional banks.

**2.2. Machine Learning**

Machine learning (ML) is another disruptive technology that enhances access to financial services. ML is a subset of AI[32], and it is a "*method of designing a sequence of actions to solve a problem, known as algorithms, which optimise automatically through experience and with limited or no human intervention.*"[33] ML can identify trends within vast, extensive datasets created from diverse and novel sources such as Internet search behaviours, viewing habits, and social media activity.[34] ML challenges traditional scoring models, which rely heavily on credit ratings based on historical payments and transactions.[35] Based on limited organised data sets, traditional credit scoring models often lead to the financial exclusion by overlooking individuals with minimal or no credit history, like young adults.[36] Conversely, ML enhances financial inclusion as it allows individuals lacking a credit history to obtain a loan or credit facility where they would conventionally have been unable to.[37] This "*self-learning capability*" of machine learning is what is truly the "*core of AI's transformative potential*".[38]

Zopa, the first P2P lending company[39], demonstrates the efficacy of machine learning in enhancing financial inclusion. Zopa launched its 'Borrowing Power' tool in 2019, which employs advanced ML methods to decipher the operations within the opaque credit scoring process. This data, combined with extra details from the client, is analysed to determine if there is potential for enhancing a portion of their credit profile and the timeframe for such improvement.[40] Another example is ZestFinance, which partnered with Baidu in 2016 to utilise the vast information Baidu had collected from its users to analyse many potential credit variables, offering loans to individuals with minimal traditional credit history in the Chinese market.[41]

---

[31] Carla Fernandes and Maria Rosa Borges and Jorge Caiado, 'The Contribution of Digital Financial Services to Financial Inclusion in Mozambique: An ARDL Model Approach' (2021) 53 AE 408.

[32] Financial Stability Board (n 19) 4.

[33] *Ibid* 4.

[34] *Ibid* 4, 9.

[35] *Ibid* 12.

[36] *Ibid* 13.

[37] *Ibid*.

[38] Carsten Maple and others, 'The AI Revolution: Opportunities and Challenges for the Finance Sector' (2023) arXiv:2308.16538 6.

[39] Zopa, 'Our Story' (2023) <https://www.zopa.com/about/our-story>, accessed on 22 December 2023.

Following the above, it is clear that alternative lenders' adoption of machine learning has innovated the credit risk assessment process. It provides hope for borrowers seeking loans previously rendered invisible or overlooked by conventional credit-scoring models. It enables alternative lenders to identify creditworthy individuals beyond the strictures of traditional credit scoring parameters. This allows alternative lenders to broaden their customer base significantly. In contrast, one should note that vigilance is necessary to mitigate the risk that these ML models may perpetuate bias or unfairness.[42] This is due to the 'black box' characteristic of ML algorithms[43] and large language models (LLMs),[44] which will be discussed later in this paper.

**2.3. Large Language Models (LLMs)**

Next, I turn to Large Language Models (LLMs). LLMs are a form of AI that have become potent instruments for various applications such as machine translation, natural language processing (NLP), and responding to queries.[45] How are LLMs trained? By handling and examining a significant volume of text, LLMs acquire an in-depth knowledge of the language, allowing them to create logical and relevant responses to the context.[46] Within the financial sector, LLMs are utilised to assess and manage risks by studying extensive previous industry historical datasets to spot potential dangers and suggest measures to counteract them using a range of financial algorithms.[47] This allows digital banks and alternative lenders to utilise LLMs to make more informed assessments of higher quality, such as more accurate credit scoring, credit approvals, and the viability of financial products.[48] Integrating LLMs into digital banks and alternative lending platforms has dramatically improved financial inclusion.

---

[40] Tamsin Fanning, 'The Tech behind Borrowing Power' (*Zopa*, 16 October 2019), <https://www.zopa.com/blog/article/the-tech-behind-borrowing-power>, accessed on 22 December 2023.

[41] Saqib Aziz and Michael Dowling, 'Machine Learning and AI for Risk Management' in Theo Lynn, John G Mooney, Pierangelo Rosati, and Mark Cummins (eds), *Disrupting Finance. Palgrave Studies in Digital Business & Enabling Technologies* (Palgrave Pivot, Cham 2019) 34.

[42] Financial Stability Board (n 19) 37.

[43] *Ibid* 13.

[44] Muhammad Usman Hadi and others, 'Large Language Models: A Comprehensive Survey of its Applications, Challenges, Limitations, and Future Prospects' (2023) TechRxiv 24.

[45] Hadi and others (n 44) 0-1.

[46] *Ibid* 9.

[47] *Ibid* 11.

[48] *Ibid* 11.

Research on LLMs is growing; however, studies are limited which cover their technical complexities and efficient use.[49]

LLMs have notably elevated the standard of customer service within the financial sector. Automated 'chatbots' and virtual assistants can use NLP to provide round-the-clock support to address customers' financial queries or promptly assist them with their transactions.[50] This support can be provided in multiple languages, enabling LLMs to reach a diverse linguistic range of communities.[51] All of this enhances the user experience of banking customers.[52] One example is the P2P lending platform LendingClub.[53] Its robo-advisor provides prudent financial support to its customers who lack experience.[54] LLMs can be a crucial service aspect for digital banks and alternative lenders lacking physical branches, giving them a leg up to compete with traditional institutional banks and surpass them by democratising financial guidance.

### 3. Regulatory Challenges Posed by New Business Models

### 3.1. Data Privacy and Security

The growth of digital banking and alternative lending highlights data privacy and security as key regulatory challenges. Based on digital platforms and analytics, these models gather extensive personal and financial data. While useful for personalised services and credit assessment, this raises concerns about data privacy breaches and misuse.[55] These major data risks are most evident from large companies such as Verizon, Amazon, and Samsung,[56] which have decided to ban their workers from utilising ChatGPT, driven by concerns about the possible risk of exposing confidential information on OpenAI's servers.[57]

One major risk of these technologies is their vulnerability to cyberattacks. As prime targets for hackers, financial institutions risk critical data breaches exposing sensitive information like bank details and credit history. Such breaches can result in significant financial liabilities and reputational damage for banks.[58] For customers, it can potentially lead to identity theft and financial fraud. For instance, the Capital One data breach in 2019 resulted in a massive data compromise of 100 million credit card users.[59] This underscores the potential severity of such cybersecurity risks.

---

[49] *Ibid* 2.

[50] Financial Stability Board (n 19) 14-5.

[51] Sameera Banu, 'Multilingual Language Models in Natural Language Processing (NLP) with Python' (*Medium*, 25 September 2023), <https://medium.com/@mail4sameera/multilingual-language-models-in-natural-language-processing-nlp-with-python-9a6d1fda4adc>, accessed on 23 December 2023.

[52] Maple and others (n 38) 12.

[53] *Ibid* 14.

[54] *Ibid*.

[55] Financial Stability Board (n 19) 10.

[56] Deloitte, 'Large Language Models (LLMs) – A Backgrounder' (2023) 4.

[57] Aaron Mok, 'Amazon, Apple, and 12 Other Major Companies that have Restricted Employees from Using ChatGPT' (*Business Insider*, 11 July, 2023) <https://www.businessinsider.in/tech/news/amazon-apple-and-12-other-major-companies-that-have-restricted-employees-from-using-chatgpt/slidelist/101676536.cms>, accessed on 20 December 2023 as cited in Deloitte (n 56) 4.

There is a real risk of breach of privacy and data protection concerns in using LLMs due to the information they are trained on.[60] For example, GPT-4 is vulnerable to "*malicious actor*[s] [which] *might inject misleading prompts, perform 'jailbreak' attacks to make the model reveal sensitive information, or use data poisoning strategies to manipulate the model's output.*"[61] There is also the possibility that LLMs "*possess the capacity to generate and disseminate misinformation or harmful content, raising concerns about the accuracy and reliability of their outputs.*"[62]

Digital banks and alternative lenders must implement robust AML/CFT measures to prevent fraud, money laundering, or illicit activities.[63] In contrast, AI technologies (e.g., neural networks) can expose fraud and illicit activities like money laundering.[64]

Regulators increasingly focus on ensuring that digital banks and lenders implement solid cybersecurity measures and data protection protocols. The EU's General Data Protection Regulation (GDPR)[65] establishes stringent data processing rules and enhances individual data control. The challenge remains in balancing user protection with encouraging innovation in the financial sector.

### 3.2. Bias and Fairness

There is a significant risk of bias and unfair and/or discriminatory outcomes that come with a reliance on machine learning algorithms.[66] Algorithms used in credit scoring and risk assessment reflect the biases of their training data. If this data is skewed or historically biased, it can lead to the machine learning algorithm or LLM perpetuating or even amplifying these biases, potentially causing discriminatory outcomes.[67] Machine learning models may still produce discriminatory results that inadvertently associate with factors such as gender, race, or religion, possibly due to geographical factors or other personal traits.[68] There is the possibility that tools based on AI and machine learning could also overlook novel kinds of risks and events, as they might excessively focus on historical occurrences during their training.[69] The same applies to LLMs, which are also not without inherent weaknesses. If the dataset includes "*biased or inaccurate information*", LLMs can potentially exhibit these biases by generating damaging results reflecting that dataset.[70]

---

[58] Maple and others (n 38) 21.

[59] Paul Bischoff, 'Financial data breaches accounted for 232 million leaked records from January 2018 to September 2023' (*Comparitech*, 4 October 2023), <https://www.comparitech.com/blog/vpn-privacy/financial-data-breaches/>, accessed on 23 December 2023.

[60] Hadi and others (n 44) 11.

[61] *Ibid* 21.

[62] *Ibid* 23.

[63] Financial Stability Board (n 19) 27.

[64] Xiao-lin Zheng and others, 'FinBrain: When Finance Meets AI 2.0' (2019) 20 FITEE 7, 914-24

[65] Trend Micro CISO Resource Centre, 'To Keep Up With Cybersecurity Laws, Go 'Federal First', (*Trend Micro*, 2 May 2023), <https://www.trendmicro.com/en_au/ciso/23/e/cybersecurity-laws-federal-first.html>, accessed on 23 December 2023.

[66] Financial Stability Board (n 19) 27.

For example, a machine learning model trained on past loan data might inadvertently learn to associate higher risks with certain demographics if those groups were historically disadvantaged or underrepresented in the credit market. This could result in unfair credit denials or higher interest rates for these groups, reinforcing existing inequalities.

Regulators are tasked with ensuring that these algorithms are transparent and fair. Legislative efforts such as the Harmonised Rules on Artificial Intelligence proposed by the European Commission[71] and the Algorithmic Accountability Act (AAA) proposed in the United States aim to compel corporate entities to assess their systems operating on machine learning algorithms for accuracy, fairness, and bias.[72] It is worth noting that neither the 116th nor the 117th Congresses passed the first two iterations of the AAA.[73] Regulating these sophisticated and often proprietary algorithms presents a significant challenge, requiring a delicate balance between promoting fairness and protecting intellectual property rights.

### 3.3. Accountability and Transparency

With the rise in automated financial decisions in digital banking and alternative lending, issues of accountability and transparency become increasingly critical. Despite their efficiency, automated machine learning systems often operate as 'black boxes' with decision-making methods that are opaque and difficult to comprehend.[74] This lack of transparency can pose problems, particularly when customers seek explanations for decisions such as credit denials or loan terms, as "*users are unable to comprehend how the system operates, makes decisions, and the underlying reasons behind those decisions.*"[75]

---

[67] Hadi and others (n 44) 17, 19.

[68] Financial Stability Board (n 19) 27.

[69] *Ibid* 25.

[70] Hadi and others (n 44) 6.

[71] Airlie Hilliard and Ayesha Gulley, 'What is the EU AI Act?' (*Holistic AI*, 12 September 2023), <https://www.holisticai.com/blog/eu-ai-act>, accessed on 23 December 2023.

[72] Senator Ron Wyden, 'S.3572 - Algorithmic Accountability Act of 2022' (Congress.Gov, 2 March 2022), <https://www.congress.gov/bill/117th-congress/senate-bill/3572>, accessed on 23 December 2023.

[73] Adam Williams, 'US Algorithmic Accountability Act: Third Time Lucky?' (*Holistic AI*, 25 October 2023), <https://www.holisticai.com/blog/us-algorithmic-accountability-act>, accessed on 23 December 2023.

[74] Financial Stability Board (n 19) 26.

The challenge for regulators is to ensure that these automated systems are accountable and that their decisions can be explained and justified. This is important for customer trust and compliance with legal standards, such as anti-discrimination laws. The challenge is compounded by the technical complexity of these systems and the proprietary nature of the algorithms used. Therefore, "appropriate circuit breakers" are necessary as a failsafe.[76]

A potential solution to tackle this problem of an absence of transparency is the development of 'explainable AI' (XAI)[77], where the decision-making processes of AI systems are designed to be interpretable and understandable by humans.[78,79] Regulators could encourage or mandate the use of XAI in the business models of digital banks and alternative lenders to enhance transparency and accountability. However, there is a lack of clarity among research experts as to what forms an adequate explanation by XAI.[80] Research studies thus far also indicate no clear consensus on how to move forward with XAI.[81] One thing remains certain. Creating explanations and improving the clarity of opaque 'black box' AI systems will boost users' trust, enabling an understanding of the logic behind the model's choices.[82] This will, in turn, spur increased mass adoption and enhance financial inclusion. One such example is the GDPR's 'right to explanation',[83] by which a user can demand clarification about a decision rendered by the algorithm that substantially affects them.[84]

## 3.4. Regulatory Lag

Technological innovation in the financial sector often outpaces regulatory development, creating a lag. Using advanced technologies, digital banks and alternative lenders frequently operate in areas with insufficient regulatory guidance, leading to uncertainties and potential risks.

---

[75] Maple and others (n 38) 19.

[76] Financial Stability Board (n 19) 26.

[77] Waddah Saeed and Christian Omlin, 'Explainable AI (XAI): A Systematic Meta-Survey of Current Challenges and Future Opportunities' (2023) 263 KBS art 110273.

[78] Maple and others (n 38) 19.

[79] Zheng and others (n 64) 921.

[80] Maple and others (n 38) 19.

[81] Saeed and Omlin (n 77) 1.

[82] Riccardo Guidotti and others, 'A Survey of Methods for Explaining Black Box Models' (2018) 51 ACM CS 5, 1-42 as cited in Saeed and Omlin (n 77) 3.

[83] Bryan Casey, Ashkon Farhangi, and Roland Vogl, 'Rethinking Explainable Machines' (2019) 34 BTLJ 1, 143-88.

[84] Bryce Goodman and Seth Flaxman, 'European Union Regulations on Algorithmic Decision-Making and a "Right to Explanation"' (2017) 38 AI M 33, 50-7 as cited in Saeed and Omlin (n 77) 2.

A significant challenge for regulators is keeping pace with these rapid technological changes. Traditional regulatory approaches, often rigid and prescriptive, struggle to accommodate the fast-evolving nature of FinTech innovations. This lag can create a regulatory vacuum, where new products and services operate without adequate oversight, potentially leading to consumer harm or systemic risks. Without adapting to these changes fast, regulators run the risk of regulatory arbitrage.

One way to address this challenge is by adopting more flexible, principle-based regulatory approaches, which set broad objectives and standards rather than prescriptive rules. This can provide the necessary flexibility to adapt to new technologies while maintaining regulatory objectives. One example is the International Organization of Securities Commissions (IOSCO),[85] which described the effects of emerging technologies, such as algorithmic trading, on market monitoring and offered suggestions for consideration, including data gathering and international collaboration.[86]

Another approach is establishing dedicated FinTech regulatory units or 'innovation hubs' within regulatory agencies.[87] These units can work closely with FinTech companies to understand emerging technologies and develop appropriate regulatory responses.[88]

Addressing these regulatory challenges requires a nuanced approach that balances the need for the protection of customers and financial stability alongside promoting growth in the financial sector. Regulators must continually adapt and evolve their frameworks to keep pace with technological advancements, ensuring that the advantages of these innovations are realised while combating possible risks.

## 4. Balancing Innovation and Regulation

As the financial sector increasingly embraces digital transformation, the need for checks and balances with effective regulation becomes much more pressing. These checks and balances are not just imposing punitive limitations or supervisory frameworks on errant banks or lenders. It involves nurturing a conducive environment where technological advancements and regulatory frameworks coexist symbiotically. How can regulators do this without stifling innovation in the FinTech space?

---

[85] Jonathan McCarthy, 'From Childish Things: The Evolving Sandbox Approach in the EU's Regulation of Financial Technology' (2023) 15 LIT1, 21.

[86] Financial Stability Board (n 19) 28.

[87] Radostina Parenti, 'Regulatory sandboxes and innovation hubs for FinTech' Study for the Committee on Economic and Monetary Affairs, Policy Department for Economic, Scientific and Quality of Life Policies, European Parliament, Luxembourg 65 (2020) 8-9, 19-20.

[88] *Ibid*.

One such example of creating a conducive environment for mutual growth is Singapore's Project Guardian, a collaborative initiative with financial industry players by the Monetary Authority of Singapore (MAS) to examine the financial benefits of the tokenisation of assets.[89]

Another approach is the implementation of regulatory sandboxes.[90,91] Regulatory sandboxes "*can allow businesses to test innovative products, services, business models and delivery mechanisms in a live environment and with proportionate regulatory requirements*."[92] These are controlled environments where new financial technologies can be tested under regulatory supervision to help firms prepare before market launch.[93,94] One such example is the UK's Financial Conduct Authority (FCA) regulatory sandbox, which allows FinTech firms to test their products in a live environment *sans* the usual regulatory repercussions.[95] Such sandboxes serve a dual purpose. They allow regulators to understand emerging technologies and their implications in real-world scenarios, enabling them to formulate regulations supporting innovation and protecting the financial system's integrity.[96] Sandboxes also encourage innovation by making it easier for FinTech companies to obtain funding in their initial development phases.[97]

Adaptive regulatory approaches are crucial in this regard. Policymakers should be flexible enough to accommodate new technologies and business models, such as those introduced by digital banks and alternative lenders, while safeguarding consumer interests and maintaining financial stability.[98] Regulators must "*consider whether FinTech-related risks are adequately dealt with under the existing regulatory framework and whether opportunities for regulatory arbitrage have opened up*."[99]

Moreover, there is a growing need for technology-agnostic but principle-based regulations. This means regulations should focus less on the specifics of technology and more on the outcomes they produce. For instance, instead of creating rules for each type of algorithm used in credit scoring, regulators could focus on the fairness, transparency, and accountability of any automated decision-making process. This shift would ensure that regulations remain relevant and effective as technologies evolve.

---

[89] Maple and others (n 38) 30.

[90] Maple and others (n 38) 31.

[91] Carney (n 7) 13.

[92] *Ibid*.

[93] *Ibid*.

[94] Maple and others (n 38) 30.

[95] Giulio Cornelli and others, 'Regulatory Sandboxes and Fintech Funding: Evidence from the UK' (2023) RF 2.

[96] *Ibid*.

[97] *Ibid*.

[98] Financial Stability Board (n 19) 32-3.

[99] *Ibid* 32.

Integrating global standards and international cooperation is crucial in FinTech, especially for digital banking and online lending platforms that often operate transnationally. International regulatory harmonisation is necessary to effectively manage risks such as money laundering, cyber threats, and cross-border fraud. Organisations such as the Financial Stability Board (FSB)[100] and the Basel Committee on Banking Supervision[101] play pivotal roles in fostering global standards and practices.[102] Collaborative efforts, such as the Global Financial Innovation Network (GFIN), facilitate dialogue and cooperation between regulators, ensuring consistent and comprehensive regulatory approaches.[103]

Regulatory policies should prioritise consumer education and protection, especially as financial services grow more digital and complex. Mandating clear guidelines and transparency ensures consumers are informed about their choices. This includes understanding digital loan terms, algorithm-driven financial advice, and data-sharing implications.

Data privacy and cybersecurity regulations are important to foster trust in digital financial services. The General Data Protection Regulation (GDPR)[104] (implemented by the European Union) is an example of a robust framework that can serve as a model. It balances the protection of individual data rights with the needs of businesses to process data for legitimate purposes.[105] Similar frameworks could be adopted globally to ensure consistent data protection standards.

Lastly, the changing landscape of financial technologies necessitates ongoing dialogue among regulators, industry professionals, academics, and consumers. Regular consultations and collaborative research are crucial to identify FinTech trends, risks, and opportunities. Continuous engagement ensures regulations stay effective and relevant, balancing innovation with strong regulatory standards.

The evolving nature of financial technology demands dynamic, adaptive regulations. Key elements include international cooperation and a focus on principle-based outcomes, crucial for maintaining the financial system's resilience. It encourages innovation that can lead to greater financial inclusion and economic growth. However, it is critical to remember that "…*a*

---

[100] Financial Stability Board, 'About the FSB' (16 November 2020) <https://www.fsb.org/about/>, accessed on 20 December 2023.

[101] Bank of International Settlements, 'Guidance on the Application of the Core Principles for Effective Banking Supervision to the Regulation and Supervision of Institutions Relevant to Financial Inclusion' (2016).

[102] Jonathan McCarthy (n 85).

[103] Global Financial Innovation Network (2023) <https://www.thegfin.com/>, accessed on 20 December 2023.

[104] European Council, 'The General Data Protection Regulation' (11 December 2023) <https://www.consilium.europa.eu/en/policies/data-protection/data-protection-regulation/>, accessed on 20 December 2023.

[105] *Ibid*.

*human in the loop is essential: we are, unlike machines, able to take into account context and use general knowledge to put AI-drawn conclusions into perspective.*"[106]

## 5. Conclusion

In conclusion, integrating digital platforms, machine learning, and LLMs within digital banks and alternative lenders represents a technological upgrade and a fundamental paradigm shift towards more inclusive financial services. These innovations have opened doors for segments of the population previously excluded from traditional financial systems. This is a significant step in making progress towards universal financial inclusion.

However, these advancements are not devoid of complications. They introduce intricate issues surrounding data privacy, security, algorithmic bias, accountability, and the pace of regulatory adaptation. This paper has examined the double-edged nature of this progression towards financial inclusion using disruptive 'enabling' technologies: their potential to democratise financial services and the ensuing regulatory challenges.

A nuanced and dynamic approach is essential, requiring regulators to balance various factors. Future research must develop frameworks that evolve with technological progress, protect consumer rights, and ensure market integrity. Policy development should promote innovation based on transparency, accountability, and fairness. International cooperation and global standards are vital to tackle the challenges of the global nature of digital financial services. A thorough re-evaluation and adaptation of regulatory frameworks are needed, demanding collaboration among policymakers, industry stakeholders, and academics.

Such collaboration is essential to fully harness the benefits of these technological advancements without compromising the financial system's stability. This balanced approach is not merely crucial for the continued growth of the financial sector. However, it is imperative to forge a more inclusive and equitable financial ecosystem.

**Acknowledgements:**

This working paper authored by Luke Lee expands upon research originally conducted as part of the academic requirements for the Master of Laws (LL.M) degree at The Dickson Poon School of Law, King's College London, during the academic year 2023. A version of this work was submitted in fulfillment of the LL.M degree. This paper aims to contribute to the scholarly discourse on the impact of digital banks, machine learning and alternative lenders on financial inclusion and the regulatory landscape. No external supervision, funding, or ethics review was required for the development of both the original submission and this expanded research.---

[106] Finextra and Intel, 'The Next Big Wave: How Financial Institutions Can Stay Ahead of the AI Revolution' (2017) as cited in Financial Stability Board (n 19) 7.

Riccardo Guidotti and others, 'A Survey of Methods for Explaining Black Box Models' (2018) 51 ACM CS 5, 1-42.

Rupert Jones, 'Zopa Exits Peer-to-Peer Lending to focus on Banking' (*The Guardian*, 11 December 2021) <https://www.theguardian.com/money/2021/dec/11/zopa-peer-to-peer-lending-p2p-money>, accessed on 20 December 2023.

Sameera Banu, 'Multilingual Language Models in Natural Language Processing (NLP) with Python' (*Medium*, 25 September 2023), <https://medium.com/@mail4sameera/multilingual-language-models-in-natural-language-processing-nlp-with-python-9a6d1fda4adc>, accessed on 23 December 2023.

Saqib Aziz and Michael Dowling, 'Machine Learning and AI for Risk Management' in Theo Lynn, John G Mooney, Pierangelo Rosati, and Mark Cummins (eds), *Disrupting Finance. Palgrave Studies in Digital Business & Enabling Technologies* (Palgrave Pivot, Cham 2019).

Senator Ron Wyden, 'S.3572 - Algorithmic Accountability Act of 2022' (*Congress.Gov*, 2 March 2022), <https://www.congress.gov/bill/117th-congress/senate-bill/3572>, accessed on 23 December 2023.

Tamsin Fanning, 'The Tech behind Borrowing Power' (*Zopa*, 16 October 2019), <https://www.zopa.com/blog/article/the-tech-behind-borrowing-power>, accessed on 22 December 2023.

The World Bank, 'Financial Inclusion' (*The World Bank*, 29 March 2022), <https://www.worldbank.org/en/topic/financialinclusion/overview>, accessed 2 December 2023.

Trend Micro CISO Resource Centre, 'To Keep Up With Cybersecurity Laws, Go 'Federal First', (*Trend Micro*, 2 May 2023), <https://www.trendmicro.com/en_au/ciso/23/e/cybersecurity-laws-federal-first.html>, accessed on 23 December 2023

Waddah Saeed and Christian Omlin, 'Explainable AI (XAI): A Systematic Meta-Survey of Current Challenges and Future Opportunities' (2023) 263 KBS art 110273.

Xiao-lin Zheng and others, 'FinBrain: When Finance Meets AI 2.0' (2019) 20 FITEE 7, 914-24.

Zopa, 'Our Story' (2023) <https://www.zopa.com/about/our-story>, accessed on 22 December 2023.17